\documentclass[runningheads]{llncs}
\usepackage[T1]{fontenc}
\usepackage{a4wide}
\usepackage[misc]{ifsym}
\usepackage{hyperref}
\usepackage{rotating}
\usepackage{graphicx}
\usepackage{xcolor}
\usepackage{amsfonts,amsmath,amssymb}
\usepackage{algorithm}
\usepackage{algpseudocode}
\usepackage{paralist}

\usepackage{svg}

\usepackage{caption}
\usepackage{subcaption}

\newcommand{\algname}{\mbox{Alpha Elimination}}

\usepackage{color}

\begin{document}
\title{Alpha Elimination: Using Deep Reinforcement Learning to Reduce Fill-In during Sparse Matrix Decomposition
}
\titlerunning{Alpha Elimination}

\author{Arpan Dasgupta\inst{1}\and
 Pawan Kumar[\Letter]\inst{1}
 \authorrunning{A. Dasgupta et al.}
 \institute{IIIT Hyderabad, India \\
 \email{arpan.dasgupta@research.iiit.ac.in} \\
 \email{pawan.kumar@iiit.ac.in} 
}
}

\toctitle{Alpha Elimination: Using Deep Reinforcement Learning to Reduce Fill-In during Sparse Matrix Decomposition}
\tocauthor{Arpan Dasgupta, Pawan Kumar}
\maketitle             

\begin{abstract}
A large number of computational and scientific methods commonly require decomposing a sparse matrix into triangular factors as LU decomposition. A common problem faced during this decomposition is that even though the given matrix may be very sparse, the decomposition may lead to a denser triangular factors due to fill-in. A significant fill-in may lead to prohibitively larger computational costs and memory requirement during decomposition as well as during the solve phase. To this end, several heuristic sparse matrix reordering methods have been proposed to reduce fill-in before the decomposition. However, finding an optimal reordering algorithm that leads to minimal fill-in during such decomposition is known to be a NP-hard problem. A reinforcement learning based approach is proposed for this problem. The sparse matrix reordering problem is formulated as a single player game. More specifically, Monte-Carlo tree search in combination with neural network is used as a decision making algorithm to search for the best move in our game. The proposed method, {\algname} is found to produce significantly lesser non-zeros in the LU decomposition as compared to existing state-of-the-art heuristic algorithms with little to no increase in overall running time of the algorithm. The code for the project will be publicly available here\footnote{\url{https://github.com/misterpawan/alphaEliminationPaper}}.

\keywords{Reinforcement Learning  \and Sparse Matrices \and Deep Learning \and LU \and MCTS}
\end{abstract}

\section{Introduction}

Computations on large matrices are an essential component of several computational and scientific applications. In most cases, these matrices are sparse which, if exploited well, provide a significant reduction in computation time and memory consumption. Therefore, it is essential to ensure that any operation performed on these sparse matrices maintains their sparsity, so that subsequent operations remain efficient.

\begin{figure}[t]
    \centering
    \includegraphics[scale=0.38]{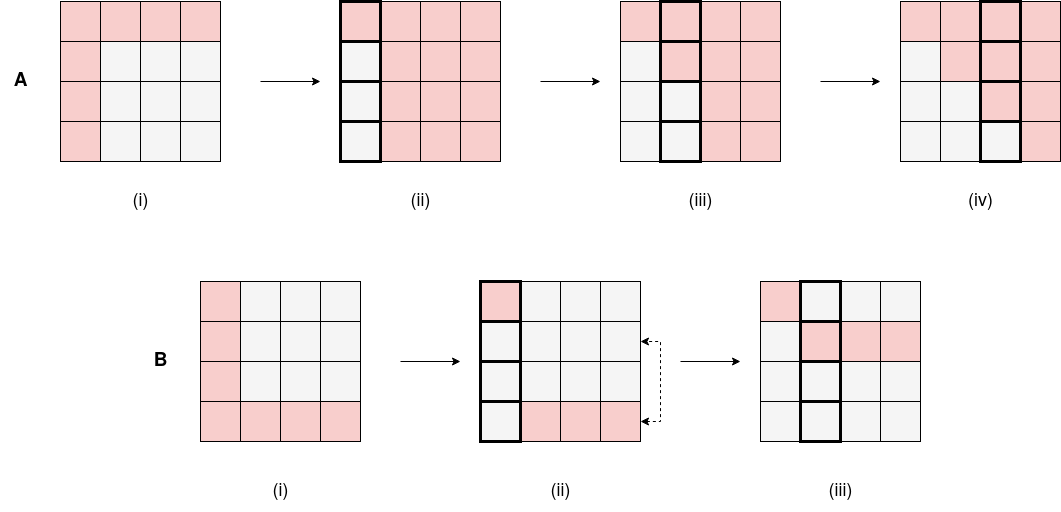}
    \caption{\label{fig:fill_in}Toy example demonstrating the effect of fill-in during LU decomposition on different permutations of the same matrix. The non-zero entries are denoted by red color and the column being eliminated is highlighted at each step. (A) The first column being eliminated causes all the other columns to become non-zeros, which are not eliminated in the subsequent steps. We end up with $10$ non-zero entries in the $U$ matrix. (B) Using a different row permutation for the matrix, we obtain no fill-in, and end up with $4$ non-zeros in the $U$ matrix.}
\end{figure}

\paragraph{\textbf{Challenges in LU decomposition for sparse matrices:}} 
One of the most frequently performed operations on matrices is decomposition. Matrix decomposition in linear algebra refers to the factorization of a matrix into a product of multiple matrices; examples include LU, QR \cite{kumar2014b,kumar2015}, SVD, etc. \cite{golub2013matrix}. Matrix decomposition is also used in preconditioning \cite{benzi2002,das2020,das2021,katyan2020,kumar2014,niu2010,kumar2013,kumar2013b,kumar2016,kumar2013c,kumar2014c}. Different types of decomposition may lead to factors with different properties or structures. In particular, in LU decomposition \cite{golub2013matrix}, the matrix $A$ is decomposed into a lower triangular and an upper triangular matrix, $L$ and $U$ respectively. The LU decomposition is useful in solving systems of linear equations stemming from numerous and wide variety of  applications ranging from 3D reconstruction in vision, fluid dynamics, electromagnetic simulations, material science models, regression in machine learning and in a variety of optimization solvers that model numerous applications. Thus it is of essence that the $L$ and $U$ matrices after the decomposition of the sparse matrix are also sparse. However, it is possible that during the matrix decomposition, the $L$ and $U$ matrices become much denser than the initial matrix due to \textit{fill-in} resulting from Gaussian elimination. 
The LU decomposition at each step requires choosing a pivot row, and then uses the diagonal entry on this row to eliminate all the elements in the same column in the following rows. The method is illustrated in Fig. \ref{fig:fill_in}. The figure shows that choosing the correct row is an important problem. The number of non-zeros created in the $U$ matrix (and similarly in the $L$ matrix) can be significantly reduced by choosing correct rows as pivot at each step.

The process of choosing a suitable row can also be formulated as finding a row permutation matrix $P_{\text{row}},$ which is pre-multiplied to the original matrix before performing the LU decomposition. The inverse of this permutation matrix can be multiplied back after the decomposition to obtain the original required decomposition without affecting the sparsity as follows
$    A = P_\text{row} ^ {-1} (P_\text{row} A) = P_\text{row} ^ {-1} (L U). $
Finding an optimal $P_{\text{row}}$ matrix which will minimize the fill-in is a very difficult task due to two reasons (1) Since the row chosen has to have a non-zero value at the column being eliminated, it is impossible to know in advance which row will have a non-zero value due to the non-zeros changing at each step of elimination (2) With larger matrices, the search space becomes impossible to traverse completely to find the optimal. In fact, \cite{ccatalyurek2011hypergraph} shows that solving this problem for a symmetric matrix is an NP-hard by converting it to a hypergraph partitioning problem.
We aim to use Reinforcement Learning as a way to find a permutation matrix, which gives a lower fill-in than existing state-of-the-art algorithms.

\paragraph{\textbf{Deep Reinforcement Learning:}} 
Deep Reinforcement Learning (DRL) has become an effective method to find an efficient strategy to games wherever a large search space is involved. Algorithms have been developed which are capable of reacting and working around their environment successfully. The success of recent RL methods can be largely attributed to the development of function approximation and representation learning methods using Deep Learning. 
A landmark success in this field came when Deep Mind developed an algorithm AlphaGo \cite{silver2016mastering}, which was able to beat the reigning Go champion using DRL. The DRL methods have since been applied to several domains such as robotics, which find control policies directly from camera input \cite{levine2016end}, \cite{levine2018learning}. RL has been used for several other applications such as playing video games \cite{mnih2015human,mehta2023,mehta2023b}, managing power consumption \cite{tesauro2007managing} and stowing objects \cite{levine2018learning}. 
One unique way to use DRL has been to aid discovery of optimal parameters in existing models such as machine translation models \cite{zoph2016neural}, and also for designing optimization functions \cite{li2016learning}. To this end, work has been recently done on algorithm discovery for matrix multiplication \cite{fawzi2022discovering} and discovery of protein structure \cite{jumper2021highly}. Our work aims to use DRL to replace heuristic algorithms for sparse matrix permutation with an algorithm which can find a much better permutation, thus significantly saving time and memory in downstream tasks, in particular for LU.

\paragraph{\textbf{Contributions:}} 

Our main contributions can be summarized as follows.
\begin{itemize} 
    \item We formulate the problem of finding the row permutation matrix for reduced fill-in as a single player game complete with state, action and rewards. We then use MCTS to find  a solution to the game.
    \item We perform extensive experiments on matrices from several real-world domains to show that our method performs better than the naive LU without reordering as well as existing heuristic reordering methods. 
\end{itemize}

\section{Related Work}

Reduction of fill-in in LU and Cholesky decomposition of sparse matrices is a well studied problem. \cite{kaya2014fill} provides an overview of the problem and describes the different algorithms in terms of a graph problem. Approximate Minimum Degree (AMD) \cite{amestoy1996approximate} is an algorithm which permutes row and columns of the matrix by choosing the node at each step which has the minimum degree in the remaining graph. This was shown to perform well on an average. Column Approximate Minimum Degree (ColAMD) \cite{davis2004algorithm} performs approximate column permutations and uses a better heuristic than AMD. SymAMD is a derivative of ColAMD which is used on symmetric matrices. Sparse Reverse Cuthill-McKee (SymRCM) \cite{liu1976comparative} is another method that is commonly used for ordering in symmetric matrices. It utilizes the reverse of the ordering produced by the Cuthill-McKee algorithm which is a graph based algorithm.
However, all these algorithms utilize heuristics with some bound on the error, but no work involving machine learning has been done in this area.

Monte Carlo tree search (MCTS), a popular Reinforcement Learning algorithm has previously proven successful in a variety of domains such as playing games like Go \cite{silver2016mastering}, Chess and Shogi \cite{silver2017mastering}. Applications have also been found in domains like qubit routing \cite{sinha2022qubit} and planning \cite{munos2014bandits} where the problems can be formulated as a single player game. Recently, \cite{fawzi2022discovering} used MCTS for discovering faster matrix multiplication algorithms. To our knowledge, there has been no attempt at application of RL in replacing heuristic algorithms for for efficient LU decomposition on sparse matrices.

\section{\algname}

This section will be organized as follows, in section one, we will talk about how we formulated the matrix decomposition problem as a single player game in the RL setting. In the next section, we will show how the Deep MCTS algorithm works in our problem, and in the third section we will talk about our choice of neural networks for the problem at hand.

\subsection{Section 1 : Formulating the Game}

\begin{figure}[t]
    \centering
    \includegraphics[scale=0.43]{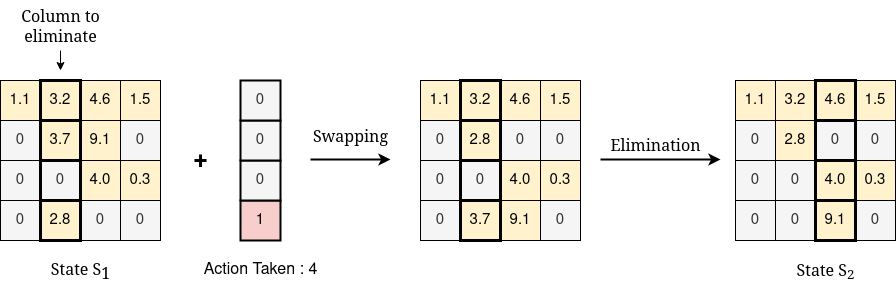}
    \caption{\label{fig:game_state} Representation of a single step of the game, starting from a state. The action chosen determines the row pivot at each point. After the swapping of the row, the elimination is performed leading to the next step. The column in bold represents the next column to be eliminated.}
\end{figure}

Fig. \ref{fig:game_state} shows the state of the game at a certain point in the elimination. 
Let us assume that the matrix is denoted by $A$ and we are at step $i$ of the elimination, i.e., we are trying to eliminate entries below diagonal of the $i_\text{th}$ column. 
Elimination of column $i$ involves (i) Picking a row $j$ as a pivot where the value at column $i$ is non-zero (ii) Swapping row $j$ for the current row $i$ (iii) Using the value $A[i][i]$, we make all values $A[k][i] = 0$ for $k > i$ by performing elementary row operations.
Fig. \ref{fig:game_state} also shows the elimination procedure.
\begin{definition} \textbf{(State)}
A state $S$ of the game provides entire configuration of the game and the environment at a certain timestep $t$. In our problem, this includes the matrix $A$ as well as the index $i$ which represents the current column to be eliminated.
\end{definition}
Although the value of $i$ could be inferred from $A$, complete state information includes both the values.

\begin{definition} \textbf{(Action)}
The action $a$ is the mechanism by which an agent transitions between the states. The agent has to choose one of several legal actions at the step. In our game, the action to be taken by the agent must be a row index $j$ where $j \ge i$ and $A[j][i] \ne 0$ or if no such value exists, $j=i$. The agent must thus choose which row to swap in for the current row as a pivot.
\end{definition} 

As we have seen in the examples, choosing this pivot correctly at each step is essential to reduce the fill-in.
The transition from one state to another is also done by using the procedure for elimination described previously.

\begin{definition} \textbf{(Reward)}
Reward $R(S, a)$ is the numerical result of taking a certain action in a state. The final goal of the agent in our game is to reduce the number of non-zeros created in total during this elimination procedure. 
\end{definition}

The reward can be provided to the agent in two different ways. We can provide the reward at the end of each step as negative of the number of non-zeros created at the current step based on the action. Alternatively, we can provide the total number of non-zeros created as a fraction of initial zeros in the matrix as a negative reward. While both these reward mechanisms are similar, the latter worked better in practice for us due to the ease of tuning the exploration factor $c$ for MCTS.

\subsection{Section 2: Applying Deep MCTS}

Monte Carlo Tree Search progresses by executing four phases in repetition: select, expand, evaluate and backup. The entire algorithm for our problem is shown in form of pseudo-code in Algorithm \ref{alg:mcts}.

\algdef{SE}[DOWHILE]{Do}{doWhile}{\algorithmicdo}[1]{\algorithmicwhile\ #1}%
\begin{algorithm}[t!]
\caption{\label{alg:mcts} Monte Carlo Tree Search Algorithm for {\algname}}
\begin{algorithmic}
\Require Starting State $S$, Model $M$, Immediate reward function $R$
\For{\textit{loop} $\gets 1$ to \textit{num\_mcts\_loops}}
    \State root $\gets S$ 
    \State cur\_node $\gets S$ 
    \While{True} 
    \Comment{Runs till expand phase reached}
        \State Select best action $A$ which minimizes UCT \quad (See \eqref{eq:uct}).  \Comment{\textbf{Select Phase}}
        \State Compute UCT using prior values and noise 
        \If{cur\_node.children[$A$] $\ne$ null}
            \Comment{Move to next state if discovered}
            \State cur\_node $\gets$ cur\_node.children[$A$]
        \Else
            \State new\_state $\gets$ cur\_state.\textbf{step}($A$)
            \Comment{\textbf{Expand Stage}}
            \If{new\_state $=$ null}
            \Comment{Leaf Node}
                \State break
            \EndIf
            \State cur\_state.children[$A$] $\gets$ new\_state
            \State \textbf{store} reward[cur\_state, $A$] 
                \State \hskip4em $\gets$ $R$(new\_state) - $R$(cur\_state)
            \State break
        \EndIf
    \EndWhile

    \State cur\_reward $\gets$ model(cur\_state)
    \Comment{\textbf{Evaluate Phase}}
    \While{cur\_node $\ne$ root}
        \State p\_action $\gets$ action from cur\_state to parent of cur\_state
        \State cur\_state $\gets$ cur\_state.parent
        \Comment{\textbf{Backup Phase}}
        \State cur\_reward $\gets$ reward[cur\_state, p\_action] + $\gamma$ cur\_reward \quad (see \eqref{eq:updates2})
        \State \textbf{Update} cur\_state.Q\_value[$A$, p\_action] with cur\_reward \quad (See \eqref{eq:updates2}, \eqref{eq:updates3})
        \State \textbf{Update} cur\_state.N\_value[$A$, p\_action] \quad (See \eqref{eq:updates1})
    \EndWhile
    
\EndFor
\end{algorithmic}
\end{algorithm}

\begin{figure}[t]
    \centering
    \label{fig:mcts}
    \includegraphics[scale=0.38]{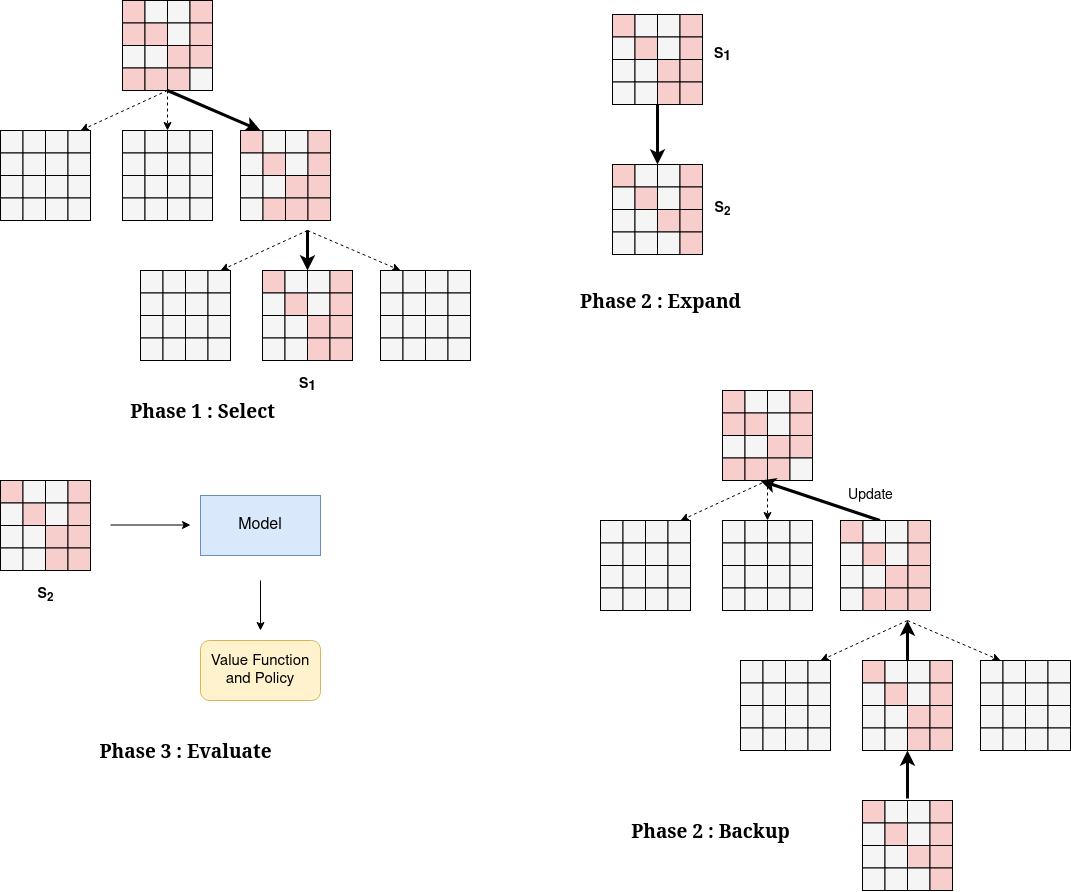}
    \caption{Four Phases of the MCTS Algorithm}
\end{figure}

\paragraph{\textbf{Select:} }
This step involves starting with a specific state and selecting a node to explore until a state node is reached which has not been explored yet. The selection starts at the root of the subtree to be explored at time $t$. The agent selects an action from a set of legal actions according to some criteria recursively.  
Assuming that $R(S, a)$ represents the immediate reward for taking action $a$ at state $S$, we keep track of two things during the MCTS procedure (over all passes during exploration):
\begin{compactenum}
    \item $N$-values: $N(S, a)$ represents the number of times the state action pair has been taken in total.
    \item $Q$-values: $Q(S, a)$ represents the expected long term reward for the state action pair $(S, a)$. To calculate this, we simply keep track of $W(S,a)$ which is the sum of all rewards received over the previous iterations. Here $N$, $W$ and $Q$ are updated as follows:
    \begin{equation}
        \label{eq:updates1}
        N(S,a) = N(S,a) + 1
    \end{equation}
    \begin{equation}
        \label{eq:updates2}
        W(S, a) = R(S, a) + \gamma W(S,a)
    \end{equation}
    \begin{equation}
        \label{eq:updates3}
        Q(S, a) = \frac{W(S,a)}{N(S,a)}. 
    \end{equation}
\end{compactenum}
These updates are actually performed in the backup stage.

We use an asymmetric formulation of Upper Confidence Bound on Trees (UCT) as a criteria for selection of the next action at each step
\begin{equation}
    \label{eq:uct}
    \text{UCT}(S,a) = Q(S,a) + c \frac{\sqrt{N(S,a)}}{N(S,a)} \times P(a|S),
\end{equation}
where $c$ represents the exploration-exploitation constant (Higher $c$ encourages exploration) and $P(a|S)$ represents the prior probability of taking action $a$ given state $S$. $P(a|S)$ is calculated by adding Dirichlet noise to the function approximator $f$ (neural network in our case) prediction. 
\begin{equation}
    P(a|S) = (1-\epsilon)f(S') + \epsilon \eta_{\alpha}. 
\end{equation}
Here, $\eta_\alpha \sim Dir(\alpha)$ where $\alpha = 0.03$, $\epsilon = 0.25$, (values are the commonly used ones described in \cite{sinha2022qubit}) and $S'$ is the resulting state when action $a$ is taken at $S$. Here $Dir(\cdot)$ stands for Dirichlet distribution. This ensures that all moves are tried while search still overrules bad moves \cite{silver2016mastering}.
The prior estimate $P(a|S)$ improves as the MCTS continues to explore. 

\paragraph{\textbf{Expand:}}
The expand step is invoked when the select reaches a state and takes an action which has not been explored yet. The step involves creating a new node and adding it to the tree structure.

\paragraph{\textbf{Evaluate:}}
On reaching a node newly created by the expand stage or reaching a leaf node, the evaluation phase is commenced. This involves estimating the long term reward for the current state using a neural network as an estimator. The neural network is provided the current state as the input, which estimates the expected reward and the prior probabilities for each action from the current state. The neural network architecture used for this purpose is described in the following section. 

\paragraph{\textbf{Backup:}}
Once the evaluation phase is completed, the reward value estimated at the last node in the tree is propagated backwards until the root is reached. At each of the ancestor nodes, the values of the expected reward for each action $Q$ and the number of times each action is taken $N$ are updated using the update equations described previously. 
As the MCTS probes the search space, it gets better estimates for the prior and the expected reward. The $N(S,a)$ represents the policies to take, and is thus used for training the policy (\ref{eq:policy}), while the average $Q$ value is used to train the value estimator (\ref{eq:value})
\begin{equation}
    \label{eq:policy}
    \pi(a|S) \propto N(S,a)
\end{equation}
\begin{equation}
    \label{eq:value}
    V(S) = \frac{\Sigma_{a}W(S,a)}{\Sigma_{a}N(S,a)}.
\end{equation}

\subsection{Section 3 : Neural Network Architecture}

The role of a neural network in the Deep MCTS algorithm is to be able to act as a function approximator which can estimate the expected reward ie. the $Q$-values for a certain state and the expected reward for each state-action pair for that state. Since actually calculating the $Q$-values is not feasible due to the intractable size of the search space, a neural network is used instead due to its ability to learn rules from the previous exploration data directly. 
\begin{figure}[t]
    \centering
    \includegraphics[scale=0.33]{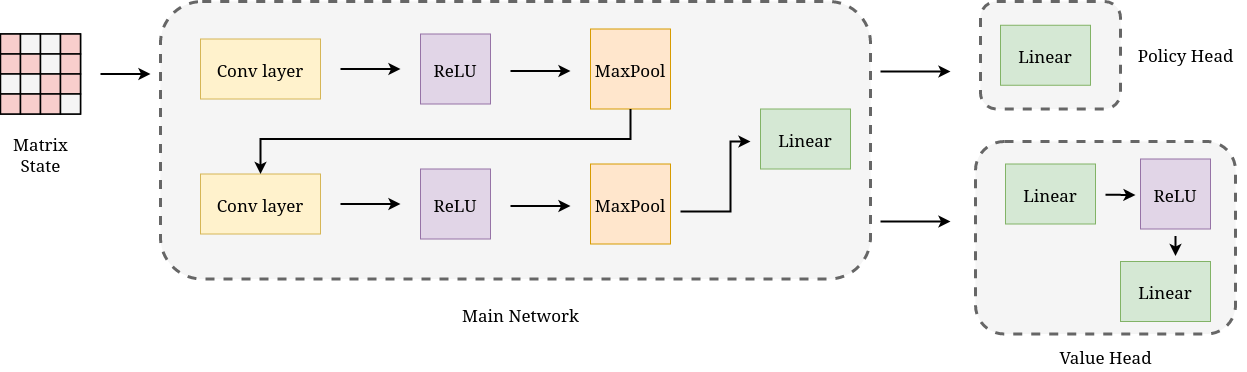}
    \caption{\label{fig:arch} Architecture of the neural network used for value and policy prediction. The first and second convolution layers have kernel sizes $3$ and $5$ respectively with a padding of $2$ and stride $1$ over $3$ channels. The MaxPool layers have a kernel size of $2$. Each of the linear layers have a single fully connected layer. The hidden layer dimensions are equal to the matrix size $N$, except for the last layers, where it corresponds to $1$ for the value head and $N$ for the policy head.}
\end{figure}
MCTS provides the neural network with a state $S$ and the network must output two values: (i) $\pi$, estimate of the $Q$ values for each action (ii) $V$, value function of that state. This is achieved by adding two output heads to the same neural network.
The main challenge at this point was on how to provide the sparse matrix as an input to a neural network. For this, we considered several possible architectures.  

Intuitively, a sparse matrix can be treated as an image with mostly zero pixels. CNNs have been successful in similar problems capturing row and column dependencies, for example, in board games such as Go \cite{silver2016mastering}, in Chess and Shogi \cite{silver2017mastering}. While sparse layers for artificial neural networks are much more efficient than CNNs, a simple method of unrolling the matrix and providing it as single-dimensional input to the network does not work as most column information is lost. For example, any two vertically adjacent elements cannot preserve that information. Thus, positional embedding would need to be added with column information. However, this approach faces scaling issues, as the parameter count is proportional to matrix size $N$. Graph neural network could also be used, but it was unclear how to provide graph structure as GNN input. This idea remains a potential future direction.

We decide to use CNN as it emphasizes the row-column relationship, and is relatively simple to understand and implement.
An issue with standard CNN is that it needs to densify the sparse matrix to be able to operate on it. Thus, we can use a sparse version of CNN \cite{spconv2022} to run the CNN algorithm on larger matrices and take advantage of the sparsity. However, the benefit in running time while using sparse convolution only comes above a certain threshold of the matrix size $N$. However, as we will see at this threshold or above of matrix size, the time taken to train the network increases significantly due to large search space and alternative ways to scale the network have to be explored anyway. 
The final network architecture is denoted in Fig. \ref{fig:arch}. 
The problem of scaling to larger matrices will be discussed in detail in section \ref{sec:scaling}.
We observed that masking of the input also led to a significant improvement in rate of learning by the model.
We discuss this in detail in section \ref{sec:masking}.

\section{{\algname}: Details}

\subsection{Training}

We train our model on randomly generated matrices with a certain sparsity level. For each train matrix, the MCTS starts the self-play at time step $0$ and generates a data point for every time step. Each data point consists of the state as input and the respective $Q$-values and updated value function as the output. These data points are added to a buffer which is used to train the neural network after a certain set of intervals. Prioritized experience replay (PER) \cite{silver2016mastering} is used to efficiently train the neural network. The use of a buffer is necessary due to the fact that each data point must be independent from the previous.

\subsubsection{Size of matrix to train:}

The size of the matrix on which we train the neural network must be fixed, since the input to the neural network cannot vary in size. However, the size of the matrix to be tested on is variable. Hence, we use the block property of LU decomposition to always change our problem into the same fixed size. For example, we train our Deep MCTS for a large matrix of size $N \times N$, but in practice we may get a matrix $A$ of size $n \times n$ where $n \le N$. We can however, convert the smaller matrix to the size it was originally trained by adding an identity block as follows
$
    \left[ 
    \begin{array}{c  c} 
      I_{N-n} & 0 \\ 
      0 & A_{n \times n}
    \end{array} 
    \right]_{N \times N}.
$
The LU decomposition of the above matrix is equivalent to that of $A$, and the row permutation matrix is also just a padded one. This procedure can also be interpreted as a intermediate time step of the LU decomposition of a larger matrix.

\subsubsection{Sparsity of matrix:}

The structure of the matrix received during testing must be similar to the ones in training, thus we need to make sure that the sparsity levels of the matrices match too. As a general rule, we see that the common matrices in practice have a sparsity level of $\ge 0.85$ (ie. more than $85\%$ of values are non-zeros). Thus, the network is trained on various levels of sparsity above this threshold.

\subsection{Prediction}

During prediction, instead of selecting the next action based on UCT, we directly utilize the output of the trained model as a prediction of the expected reward. We therefore choose optimal action $a^{*} = \max_{a} \pi(a|S)$ where $\pi$ is output of policy function of neural network. 
At each step, the algorithm outputs the row $j$ to be swapped with current row $i$. Using this information, we can reconstruct the row permutation matrix $P$ for final comparison with the other methods.

\subsection{\label{sec:masking} Remark on the Role of Masking}

Neural networks are powerful learners; however, when presented with an abundance of information, they may struggle to discern the essential aspects of a problem and might establish spurious correlations. In our experiments, we observed that providing the original matrix with floating point entries to the neural network resulted in slow and erratic learning. Consequently, we employed masking of the non-zeros to minimize noise and ensure that the neural network focuses on the role of non-zero entries as pivots.
More specifically, the non-zero entries are converted to a constant value of $1$, whereas, the zeros (accounting for floating-point errors) are assigned a value of $0$. This masking technique assumes that none of the non-zeros will become zero during elimination. Although this assumption may not be universally true, it is valid in most cases involving real-world matrices with floating-point values.
Even in cases where the assumption does not hold, such instances are relatively rare and have minimal impact on the overall policy. Masking also has negligible effect on time complexity.

\subsection{\label{sec:scaling} Scaling to Larger Matrices}

The method effectively finds row permutation matrices, but the search space for even small matrices is vast. Matrix size $N$ and training data requirements increase with larger matrices. In real-life applications, matrix sizes can reach up to millions or higher. To address this, we employ a graph partitioning algorithm from the METIS library \cite{karypis1997metis} to partition the matrix into parts of size $500$, which allows for efficient learning within a reasonable time-frame. We remark here that most LU factorization for large matrices are anyway partitioned into small parts to achieve parallelism on modern day multi-core or multi-CPU architectures \cite{golub2013matrix}, and only ``local'' LU factorization of the smaller sub-matrices are required. 

\section{Experiments and Discussion}

\subsection{Experimental Setup}
 
Experiments were conducted on a Linux machine with 20 {\tt Intel(R) Xeon(R) CPU E5-2640} v4 cores @ 2.40GHz, 120GB RAM, and 2 {\tt RTX 2080Ti} GPUs. The total number of non-zeros in the LU decomposition is used as the evaluation metric, as our method aims to minimize it. We compared our approach to the naive LU decomposition in sparse matrices and existing heuristic algorithms that minimize fill-in, such as ColAMD \cite{davis2004algorithm}, SymRCM \cite{liu1976comparative}, and SymAMD. There are some specific re-ordering techniques, but due to lack of general applicability we do not compare with them. After exporting the matrix to MATLAB, where these methods are implemented, LU decomposition was performed.
The final evaluation involved matrices from the SuiteSparse Matrix Collection \cite{kolodziej2019suitesparse}. Table \ref{table:matrices} displays the selected matrices, which span various application areas, symmetry patterns, and sizes ranging from 400 to 11 million elements.

\subsection{Experimental Results} 

\begin{table}
 \centering
\caption{\label{table:matrices} Matrices used for testing from Suite Sparse Matrix Market}
 \bgroup
 \setlength{\tabcolsep}{1.5pt}

\begin{tabular}{|p{0.9in}|p{2in}|p{0.7in}
|p{0.85in}|}
\hline
\textbf{Matrix} & \textbf{Domain} & \textbf{Rows $N$} & 
\textbf{Structurally
Symmetric (Y/N)} \\
\hline
west0479 & Chemical Process Simulation \newline Problem & 479 & 
Yes \\
mbeause	& Economic Problem & 496	& 
No  \\
tomography & Computer Graphics / Vision \newline Problem & 500 & 
No  \\
Trefethen\_500 & Combinatorial Problem  & 500	& 
Yes \\
olm500 & Computational Fluid Dynamics \newline Problem & 500 & 
Yes \\
Erdos991 & Undirected Graph & 492 & 
Yes \\
rbsb480 & Robotics Problem & 480 & 
No \\
ex27 & Computational Fluid Dynamics \newline Problem & 974 & 
No \\
m\_t1 & Structural Problem & 97,578 & 
Yes  \\
Emilia\_923 & Structural Problem & 923,136 & 
Yes  \\
tx2010 & Undirected Weighted Graph & 914,231& 
Yes \\
boneS10 & Model Reduction Problem & 914,898 & 
No \\
PFlow\_742 & 2D/3D Problem & 742,793 & 
Yes  \\
Hardesty1 & Computer Graphics / Vision \newline Problem & 938,905 & Yes \\
vas\_stokes\_4M & Semiconductor Process  Problem & 4,382,246 & No  \\
stokes & Semiconductor Process Problem & 11,449,533 & No \\
\hline
\end{tabular}
 \egroup
\normalsize
\end{table}

{
\begin{table}
\centering

\caption{\label{table:results} Total Non-Zero Count in Lower and Upper Triangular Factors after re-orderings for matrices from Suite Sparse Matrix Market Dataset. }
 \bgroup
 \setlength{\tabcolsep}{1.5pt}
\begin{tabular}{|p{0.9in}|p{0.75in}|p{0.75in}|p{0.75in}|p{0.75in}|p{0.75in}|}
\hline
\textbf{Matrix} & \multicolumn{5}{c|}{\textbf{LU Methods}} \\
\cline{2-6}
    & \textbf{Naive LU}     &\textbf{ColAMD}    &\textbf{SymAMD}    &\textbf{SymRCM}    &\textbf{Proposed \newline Method}\\
\hline
west0479 & 16358 & 4475 & 4510 & 4352 & \textbf{3592} \\
mbeause	& 166577 & 126077 & NA & NA & \textbf{94859} \\
tomography & 108444 & 41982 & NA & NA & \textbf{35690} \\
Trefethen\_500 & 169618 & 150344 & 153170 & 119672 & \textbf{94632} \\
olm500 & 3984 & \textbf{3070} & \textbf{3070} & \textbf{3070} & \textbf{3070}\\
Erdos991 & 61857 & 4255 & 4287 & 4372 & \textbf{3584} \\
rbsb480 & 192928 & 63783 & NA & NA & \textbf{55185} \\
ex27 & 122464 & 104292 & NA & NA & \textbf{63948} \\
m\_t1 & 9789931 & 9318461 & 8540363 & 8185236 & \textbf{7398266} \\
Emilia\_923 & 5.67E08 & 4.49E08 & 4.29E08 & 4.56E08 & \textbf{3.9E08} \\
tx2010 & 1.48E10 & 3.83E09 & 2.34E09 & 2.44E09 & \textbf{1.3E09} \\
boneS10 & 3.98E08 & 1.89E08 & NA & NA & \textbf{1.1E08}\\
PFlow\_742 & 1.98E08 & 9.20E07 & 8.43E07 & 8.92E07 & \textbf{8.3E07} \\
Hardesty1 & 6.03E08 & 5.91E08 & 5.90E08 & 5.92E08 & \textbf{4.9E08}\\
vas\_stokes\_4M & 1.35E09 & 8.71E+08 & NA & NA & \textbf{5.9E08} \\
stokes & 9.78E10 & 6.42E10 & NA & NA & \textbf{3.9E10} \\
\hline
\end{tabular}
 \egroup
\end{table}
\normalsize
}

\subsubsection{Comparison of methods:}

 The comparison between {\algname} and the baseline as well as the other methods is shown in Table \ref{table:results}. As it is evident from the results, {\algname} obtains significant reduction in the number of non-zeros as compared to the other methods. This leads to significant reduction in  storage space for the factors of the sparse matrices, and leads to reduction in solve time using LU factorization. 
The reduction in the number of non-zeros provides even more significant memory savings when the size of the matrices increases. Our method produced up to $61.5\%$ less non-zeros on large matrices than the naive method and up to $39.9\%$ less non-zeros than the best heuristic methods. While in some matrices our method gives a significant reduction, some matrices are much simpler in structure, providing much lesser time for improvement over simple algorithms. For example, the matrix \textit{ohm500} has a very simple structure (almost already diagonal) and it is trivial for every row or column reordering algorithm to figure out the optimal ordering. Thus, all the methods end up having the same number of non-zeros.
Some of these fill-reducing ordering methods are not applicable for non-symmetric matrices, hence applied on symmetric part $A + A^T$ of a matrix $A$; whereas, our proposed method is not restricted by structural assumptions on the matrix.

\subsubsection{Time Comparison:}

Table \ref{table:time_total} presents the total time taken for finding the permutation matrix and subsequently performing LU decomposition. The time required for LU decomposition decreases when the algorithm processes fewer non-zeros. As shown in Table \ref{table:time_lu}, the time consumed for performing LU decomposition after reordering is proportional to the number of non-zeros generated during the decomposition process.
For smaller matrices, the time saved during LU decomposition is overshadowed by the time required for ordering. However, with larger matrices, our method not only achieves a reduction in the number of non-zeros but also results in a noticeable decrease in LU decomposition time.

{
\begin{table}[t!]
\centering

\caption{\label{table:time_lu} Comparison of time (in seconds) taken for LU factorization after reordering by different methods. 
}
 \bgroup
 \setlength{\tabcolsep}{1.5pt}
\begin{tabular}{|p{0.85in}|p{0.70in}|p{0.70in}|p{0.70in}|p{0.70in}|p{0.70in}|}
\hline
\textbf{Matrix} & \multicolumn{5}{c|}{\textbf{Time taken for LU (s)}} \\
\cline{2-6}
    & \textbf{Naive LU}     &\textbf{ColAMD}    &\textbf{SymAMD}    &\textbf{SymRCM}    &\textbf{Proposed \newline Method}\\
\hline
mbeause	 & 0.0326 & 0.0319 & NA & NA & \textbf{0.0302} \\
tomography  & 0.04250 & 0.0394 & NA & NA & \textbf{0.0286} \\
Trefethen\_500  & 0.0498 & 0.0419 & 0.0392 & 0.0347 & \textbf{0.0302} \\
m\_t1  & 5.8790 & 4.7894 & 4.1321 & 3.7031 & \textbf{3.2820} \\
tx2010  & 24.3018 & 15.7040 & 14.5840 & 15.6194 & \textbf{12.9365} \\
\hline
\end{tabular}
 \egroup
\end{table}
\normalsize
}
{
\begin{table}[t!]
\centering

\caption{\label{table:time_total} Comparison of total time (in seconds) for LU (including reordering). 
}
 \bgroup
 \setlength{\tabcolsep}{1.5pt}
\begin{tabular}{|p{0.85in}|p{0.70in}|p{0.70in}|p{0.70in}|p{0.70in}|p{0.70in}|}
\hline
\textbf{Matrix} & \multicolumn{5}{c|}{\textbf{Time taken for LU (s)}} \\
\cline{2-6}
    & \textbf{Naive LU}     &\textbf{ColAMD}    &\textbf{SymAMD}    &\textbf{SymRCM}    &\textbf{Proposed \newline Method}\\
\hline
mbeause	 & \textbf{0.0326} & 0.0345 & NA & NA & 0.0336 \\
tomography  & 0.0425 & 0.0475 & NA & NA & \textbf{0.0391} \\
Trefethen\_500  & 0.0498 & 0.0437 & 0.0404 & 0.0362 & \textbf{0.0334} \\
m\_t1 & 5.8790 & 5.2174 & 4.5281 & 4.1161 & \textbf{3.9320} \\
tx2010  & 24.3018 & 16.2510 & 15.077 & 16.0324 & \textbf{13.7185} \\
\hline
\end{tabular}
 \egroup
\end{table}
\normalsize
}

\subsubsection{Hyperparameter Tuning and Ablation Study:}
The training is stopped when the average reward no longer improves or the average loss does not decrease. As a standard, either of these conditions were generally met for $N=500$ size matrices by iteration $300$. The time taken for training matrices of size $10$, $50$, $100$, $250$, $500$ and $1000$ is $0.09$, $1.2$, $6.3$, $13.1$, $27.7$ and $122.5$ hours respectively for $100$ iterations. The number of iterations also increases with increase in $N$.

\begin{figure}[t!]
    \centering
    \begin{subfigure}{.5\textwidth}
        \includegraphics[scale=0.48]{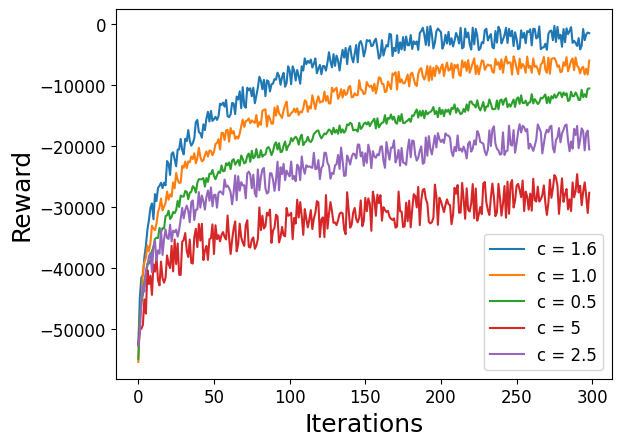}
        \caption{\label{fig:expl_graph}}
    \end{subfigure}%
    \begin{subfigure}{.5\textwidth}
        \includegraphics[scale=0.48]{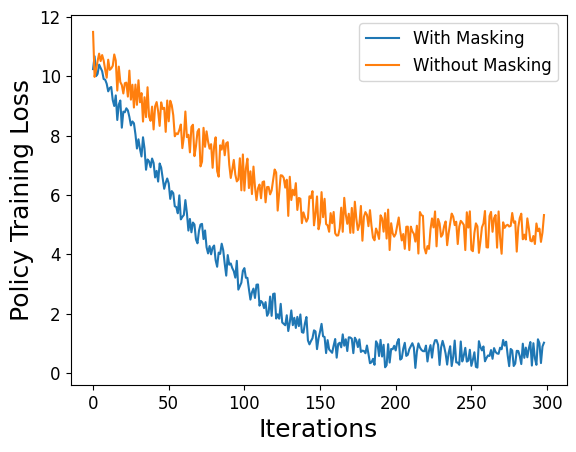}
        \caption{\label{fig:graph_mask}}
    \end{subfigure}
    \caption{(a) Reward plot versus iterations for different exploration factors $c$. (b) Training loss vs Iterations for masked and non-masked input. 
    }
\end{figure}

The most difficult hyperparameter to train is the exploration factor $c$. The correct value of $c$ determines how quickly the MCTS finds better solutions and how much it exploits those solutions to find better ones. This value is found experimentally. This is best demonstrated using Fig. \ref{fig:expl_graph}. The advantage of masking the matrix before providing it as input is demonstrated in Fig. \ref{fig:graph_mask}. The graph shows that masking helps the neural network learn better.

\section{Conclusion and Future Work}

In this paper, we demonstrated that the problem of identifying row permutations to minimize fill-in during LU decomposition of sparse matrices can be effectively formulated as a single-player game. Monte Carlo Tree Search combined with a deep neural network proves to be the optimal approach for addressing this problem.
The neural network employed for the sparse matrix serves as a critical component of the algorithm. Further research focusing on the development of scalable architectures capable of handling large sparse matrix inputs may enhance the quality of the policy output. A combination of the heuristic methods along with MCTS for bootstrapping with additional training data can be explored in the future to get further improvements. For the stability of LU, the largest pivot is generally brought to the diagonal. Our method does not always follow this and how to improve numerical stability is left as a future research direction.
Moreover, reinforcement learning methods can be potentially employed to either replace or improve upon existing heuristic algorithms, opening up new avenues for future investigation.

\section{Acknowledgment}

This work was done at IIIT-HYDERABAD, India. We thank the institute for HPC resources. We also thank Qualcomm Faculty Award (2022).

\clearpage

\section*{Ethical Considerations}
This work concerns algorithm development for sparse matrix factorization. 
To the best of our knowledge, we declare that there are no immediate or far reaching ethical 
considerations.

\bibliographystyle{splncs04}
\bibliography{AlphaElim}

\end{document}